\documentclass[10pt, a4paper]{article}

\usepackage{lrec2026} % this is the new style
\usepackage{subfig}

\title{The In-Car Sign Language Corpus (ICSL): A Multi-Modal Resource for Constrained-Space Sign Language Recognition}

\name{Raviteja Boddu$^*$, Guilherme Vieira Leite$^\dagger$, Joed Lopes da Silva$^*$, \\
{\bf\large Ângelo Benetti$^\dagger$, Isabela Barbieri$^\dagger$, Natália de Melo Afonso$^\dagger$,} \\
{\bf\large Thyago Santos$^\ddagger$, Helio Pedrini$^\dagger$, Felipe Venâncio Barbosa$^\ddagger$,} \\ 
{\bf\large  José Mario De Martino$^\dagger$, Munir Georges$^*$, Alessandro Zimmer$^*$}}

\address{$^*$Technische Hochschule Ingolstadt (THI), Bayern, Germany \\
         $^\dagger$Universidade Estadual de Campinas (UNICAMP), São Paulo, Brazil \\
         $^\ddagger$Universidade de São Paulo (USP), São Paulo, Brazil \\
         $^*$\{raviteja.boddu, joed.lopesdasilva, munir.georges, alessandro.zimmer\}@thi.de\\
         $^\dagger$martino@unicamp.br, $^\dagger$guilherme.leite@ic.unicamp.br}

\abstract{
This paper addresses the challenges of using sign language 
within shared mobility services, such as taxis, carpools, or ride-sharing platforms. The use of sign language recognition (SLR) in real-world, confined environments, specifically vehicle interiors remains largely unexplored. To motivate research in this area, we present the In-Car Sign Language (ICSL) dataset for Brazilian Sign Language (Libras), with the long-term goal of improving public transport accessibility for the Deaf and Hard-of-Hearing community.
The dataset consists of: 
(1) high-precision laboratory motion capture (MoCap) data to establish an idealized linguistic baseline and 
(2) real-world multi-modal in-car recordings captured using a 2D camera and 3D Time-of-Flight sensors.
The dataset provides a basis for comparative analyses between synthesized signing avatar animations and recorded real signing interpreter videos, which enable future research into robust ``in-the-wild'' SLR models and domain adaptation.
We describe in detail the use cases, the setup, the data collection protocol, and the metadata structure of the corpus. In total, we recorded a multimodal dataset exceeding 1.5 million frames, comprising the synchronized
multimodal streams described above featuring Libras users across various in-car scenarios. The corpus is provided with gloss annotation of lexical signs and non-lexical sign language elements specially designed to support the training and evaluation of deep neural networks for constrained space recognition.
In-vehicle signing offers a technically significant example of a constrained, occluded, and non-frontal environment. While recognizing the diverse communication strategies already employed by the Deaf community, identifying automotive-specific limitations provides a useful stepping stone for research into enhancing in-car accessibility and passenger quality of life.
\\ \newline  }

\begin{document}

\tolerance=999
\sloppy

\maketitleabstract

\section{Introduction}
\vspace{-0.5cm}
% {\color{blue}translation in introduction somewhere}
The advancement in natural language processing and computer vision has brought us closer to seamless human-machine interaction. %However, the Deaf and Hard-of-Hearing (DHH) community remains significantly underserved in ``on-the-go'' scenarios.
%% One of the most critical yet under-researched environments for accessibility is the interior of vehicles within shared mobility services, such as taxis and ride-sharing platforms.
%One of these scenarios is the interior of vehicles within shared mobility services, such as taxis and ride-sharing platforms.
While the Deaf and Hard-of-Hearing (DHH) community has developed effective strategies for "on-the-go" communication, such as the use of mobile devices or visual cues, the interior of vehicles within shared mobility services represents a technically significant and under-researched environment for Sign Language Recognition (SLR).

Specifically, the car cabin serves as a critical case study for constrained, occluded, and non-frontal signing environments, which are often overlooked in traditional laboratory-based datasets.
While Brazilian Sign Language (Libras) has an established linguistic foundation, there is currently no specialized resource documenting how signing is produced and processed in the highly constrained physical and visual environments of a car cabin~\cite{santos2026properbodylandmarksubset,Peiris2025ExploringOneHandedSigning}.

The relevant studies and corpora that exist for Libras are predominantly recorded in laboratory-controlled settings with optimal lighting and neutral backgrounds. In such environments, signers have a full, unobstructed signing space~\cite{santos2026properbodylandmarksubset}. However, the interior of a vehicle introduces severe constraints, such as seatbelts cutting across the torso, occlusions, dashboard obstruction, and constrained space that limit the range of torso, elbow, and arm movements, and dynamic environmental lighting creates substantial visual noise. These factors represent fundamental challenges for current SLR systems, which often fail to generalize to such ``in-the-wild'' conditions.

\begin{figure*}[t!] 
    \centering
    \includegraphics[width=\textwidth, height=10cm, keepaspectratio]{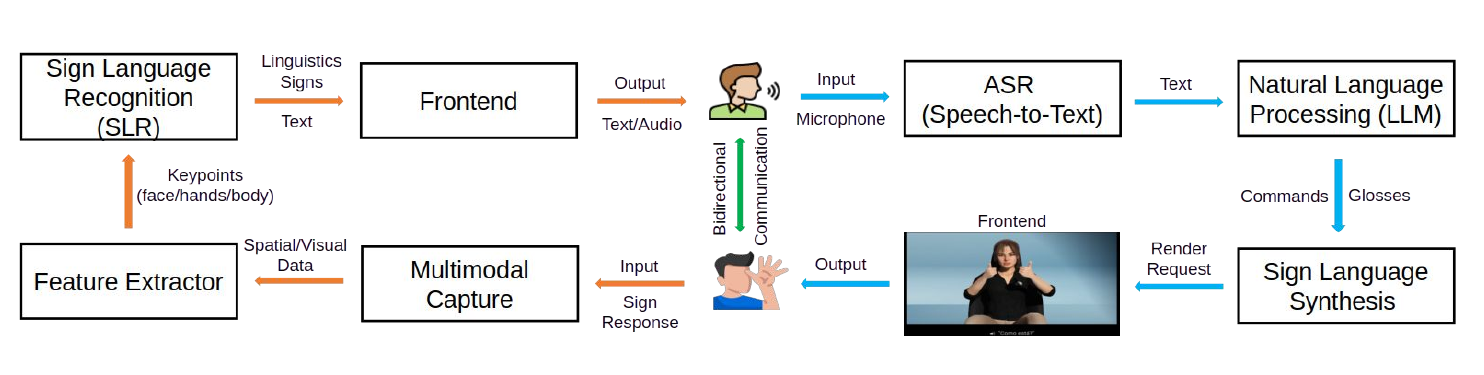} 
    \caption{Proposed Project UNITY architecture for a conceptual bidirectional loop: (right) Driver-to-Passenger Speech-to-Sign Synthesis  and (left) Passenger-to-Driver Sign Language Recognition.}
    \label{fig:architecture_top}
\end{figure*}

Within the framework of Project UNITY, a collaborative research initiative between the
%Universidade Estadual de Campinas - 
UNICAMP (Brazil) and the 
%Technische Hochschule Ingolstadt - 
THI (Germany), we seek to establish a specialized multimodal dataset
%(image/video, point cloud, IR, depth, MoCap, and textual annotation) 
that captures real-world constraints while providing a high-fidelity baseline. 
%Our objective is to facilitate the development of robust SLR models by comparing real in-car video with synthetic signing avatars.
We aim to enhance the robustness of SLR models by utilizing signing avatars as a controlled reference to evaluate the environmental challenges present in real-world in-car video.
%In this data acquisition, which is led by professional Libras signers and Deaf researchers, we captured the authentic challenges of the transport domain. We recorded a real-world dataset inside a car, focusing on passenger-driver communication scenarios. 
Led by professional Libras signers and Deaf researchers, who are also co-authors of this study, the data acquisition process seeks to document the authentic challenges of the constrained transport domain by documenting passenger-driver communication within the vehicle cabin.

The long-term vision of this project is to investigate the feasibility of a real-time, bidirectional translation framework, as seen in Figure~\ref{fig:architecture_top}, which serves as the target architecture for our ongoing work.
%The overarching goal of this project is to bridge this communication gap through a real-time, bidirectional translation framework as seen in Figure~\ref{fig:architecture_top}. 
Unlike traditional one-way approaches, our target architecture facilitates a conversational loop between driver speech and passenger signs (Libras). Building such a system requires resources that capture both idealized linguistic baselines and the severe physical restrictions of the vehicle cabin.
%Unlike previous one-way approaches that focus solely on translating driver instructions to the passenger, our proposed architecture facilitates a complete conversational loop between spoken language used by the driver and sign language (Libras) used by the DHH passenger. However, building such a system requires a high-quality multimodal resource that captures both idealized linguistic baselines and the severe physical constraints of the vehicle cabin.

% Rather than presenting a final recognition solution, this paper describes first step towards the In-Car Sign Language (ICSL) corpus which serves as the essential foundation and protocol required to guide more complex, large-scale data collection efforts in the automotive domain. It details the acquisition protocol, use-case design, and setup for in-car recordings; the technical setup of the laboratory MoCap baseline; and the architecture of the signing avatar pipeline. By providing this comparative resource, we aim to support future studies on robust, accessible communication technologies for the DHH community in transportation systems.

Rather than presenting a final recognition solution, this paper describes an initial step toward the In-Car Sign Language (ICSL) corpus, which we hope can serve as a helpful reference for future data collection efforts in the automotive domain. It details the acquisition protocol, use-case design, and setup for in-car recordings; the technical setup of the laboratory MoCap baseline; and the architecture of the signing avatar pipeline. By providing this comparative resource, we hope to contribute to future studies on accessible communication technologies for the DHH community in transportation systems.

\vspace{-0.6cm}
\section{Related Work}
\vspace{-0.2cm}
For the Brazilian Sign Language (Libras) community, there is a big gap between lab research and real-world use, especially in ride-hailing and ride-sharing services. The space, movement, and social setting within a vehicle influence how signs are produced. This section provides an overview of existing language resources, the algorithms used, and the real-world challenges involved in building a Libras dataset and tools designed specifically for use in vehicles.
\vspace{-0.2cm}
\subsection{The Landscape of Brazilian Sign Language (Libras) Resources}
\vspace{-0.2cm}
The development of language resources for Libras has changed over time. It started mainly with written documentation and later moved to video-based, multimodal data. Libras is recognized as a legal means of communication and expression in Brazil, with its own grammar and linguistic system for transmitting ideas and facts~\cite{presidencia2002_10436}.

\vspace{0.2cm}
Early work on documenting Libras mostly focused on creating dictionaries and bilingual corpora to support education and translation~\citelanguageresource{quadros2021corpuslibrasacre}. A comparative summary of prominent Libras datasets and their recording environments is provided in Table~\ref{tab:libras_datasets}.
\vspace{-0.2cm}
\subsubsection{Textual and Bilingual Corpora}
\vspace{-1mm}
One of the important recent resources for Libras is VLibras-DB~\citelanguageresource{VLibrasBD}. It is a bilingual text dataset with about 127,000 translation pairs between Brazilian Portuguese and Libras. This dataset was mainly created to help neural machine translation systems convert written text into glosses. %Gloss is a written form used to represent signs and works like a bridge between spoken and sign languages. 
Broadly defined, glossing is the use of written language to represent primary data, whether signed or spoken. Within SLR, it serves as a symbolic bridge, utilizing labels from a spoken language to represent the manual and non-manual components of a sign. 

\vspace{0.2cm}
To build VLibras-DB, a group of 10 professional interpreters worked together to define annotation rules for representing signs using glosses that
%could benefit neural network training. 
could leverage neural network training.
The gloss convention established includes directional verbs, intensifiers, and negation. The dataset primarily covers manual lexical signs but does not address challenges posed by non-manual markers or timing-related information. Non-manual markers (NMMs) are linguistically meaningful articulations produced by the face, head, torso, and gaze used as part of a sign language's phonology, morphology, syntax, and prosody~\cite{Herrmann14}. As a result, the VLibras-DB is not fully suitable for sign recognition tasks in real-world settings.

%include visual or temporal information. Because of that, it is not suitable for sign recognition tasks in real-life situations. Also, gloss is only a simplified written representation, so it cannot fully capture the physical details and variations that happen when people actually produce signs~\citelanguageresource{VLibrasBD}.

\begin{table*}[!htb]
\setlength{\tabcolsep}{2.4mm}
\centering
\caption{Overview of existing and proposed Brazilian Sign Language (Libras) datasets.}
\label{tab:libras_datasets}
\small 
% This redefines X to be vertically centered (middle) and left-aligned
\renewcommand\tabularxcolumn[1]{m{#1}}
\newcolumntype{L}{>{\raggedright\arraybackslash}X}
% Fixed-width column for columns with less text to save space
\newcolumntype{P}[1]{>{\raggedright\arraybackslash}m{#1}}

\begin{tabularx}{\textwidth}{P{2.3cm} L L c L L}
\toprule
\textbf{Dataset Name} & \textbf{Modality} & \textbf{Primary Task} & \textbf{Language} & \textbf{Scope} & \textbf{Annotations} \\ \hline
VLibrasBD & Text/Gloss & Neural Machine Translation & Libras/BP & 127k Pairs & Gloss, Syntax \\ \hline
MINDS-Libras & RGB Video & Isolated \newline Sign Recognition & Libras & 1,200+ Samples & Gloss \\ \hline
LIBRAS-UFOP & RGB Video & Isolated \newline Sign Recognition & Libras & 2,000+ Samples & Gloss \\ \hline
LIBRAS-UFSC & Video/Bank & Linguistic Documentation & Libras & National Inventory & Phonetics \\ \hline
%SignBD-Word-90 & RGB/Skeleton & Word-level SLR & Libras & 90 Classes & MediaPipe \\ \hline
LSWH100 & Synthetic RGB & Handshape Classification & Libras & 144k Images & Keypoints \\ \hline
\textbf{Ours (ICSL)} & \textbf{RGB/Depth/IR /Point Cloud} & \textbf{Constrained \newline Sign Recognition} & \textbf{Libras} & \textbf{1.5M+ Frames, 127 Phrases*} & \textbf{Gloss} \\ \bottomrule
\multicolumn{6}{l}{\footnotesize{*Current subset of a total 1,344 defined phrases.}} \\ 
\end{tabularx}
\end{table*}

%\vspace{-0.2cm}
\subsubsection{Visual and Isolated Sign Recognition Benchmarks}
\vspace{-1mm}
For isolated sign language recognition (ISLR), researchers typically use datasets such as MINDS-Libras~\citelanguageresource{MINDS-Libras} and LIBRAS-UFOP~\citelanguageresource{LIBRAS-UFOP}. These datasets have videos of single signs, mostly recorded in a controlled setup, with a plain background and the signer facing the camera. 
%Recently, researchers have improved results by using skeleton-based image representations together with 2D CNN models. With this approach, they reached an accuracy of up to 0.93 on MINDS-Libras. 
%The idea is to take movement information from the video-like body pose, hand positions, and facial landmarks and convert it into one image. 
Recently, researchers have improved recognition performance by utilizing skeleton-based image representations paired with 2D CNN models, achieving accuracies up to 0.93 on MINDS-Libras. This technique involves transforming skeletal sequences comprising body, hand, and facial landmarks into a single spatio-temporal image, allowing a 2D CNN to recognize the sign from this unified input.
This image then goes into a classifier to recognize the sign~\cite{alves2024enhancingbraziliansignlanguage}. 
%But these datasets do not really work well for car environments. 
However, such datasets do not work well for car environment. Most ISLR benchmarks assume the signer is facing the camera directly, which is very different from the angles you get inside a vehicle~\citelanguageresource{3DLEXv1}. While vehicle interiors present unique spatial challenges, they share commonalities with other atypical or constrained-space signing environments. For instance, the PopSign dataset~\citelanguageresource{PopSign_NEURIPS2023} explores the complexities of one-handed signing with the narrow field of view  of a smartphone camera. Much like PopSign, our ICSL dataset addresses the problem of reduced signing volume, where the signer must adapt the scale of their gestures due to spatial boundaries. Our work extends this by focusing on the specific two-handed constraints, such as seatbelt occlusions and non-frontal camera perspectives, unique to the automotive cabin that are not present in handheld mobile datasets.

\vspace{-0.3cm}
\subsubsection{Regional and Documentary Inventories}
The LIBRAS-UFSC corpus and the Libras National Inventory are focused on preserving Brazil's Deaf heritage and the language's diversity. They ensure the corpus includes people of different ages, genders, and regions, using careful methods to build it. The LIBRAS-UFSC corpus provides phonetic and phonological details, inspired by the Dutch Signbank model, and classifies signs by hand shape, movement, and location. These resources are relevant for linguistic research, but they are usually recorded in controlled settings, so they do not capture challenges such as visual noise, occlusions, or restricted movement that you would see in a car~\citelanguageresource{quadros2021corpuslibrasacre}~\citelanguageresource{LSWH100}.

\vspace{-0.2cm}
\subsection{Sign Language Recognition Architectures and Methodologies}
The technological evolution of SLR has transitioned from traditional manual feature extraction to end-to-end deep learning models. This shift is critical for handling the high variability of sign language, which is based on precise hand movements, facial expressions, and body language.

% \textcolor{red}{we could add our dataset info at the end of the table...if we could justify all column parameters}
\vspace{-0.2cm}
\subsubsection{Feature Extraction and Temporal Modeling}
Feature extraction focuses on capturing spatial information about hand shape, position, and orientation, as well as facial landmarks. Temporal modeling addresses the multichannel sign language characteristics (hands, facial expression, gaze, torso shift, head movement) presented in continuous stream of video frames~\cite{Camgoz20,Rastgoo21,Allyami24,Taher25}. Common architectures include:

\begin{itemize}
    \item \textbf{CNN-BiLSTM:} Convolutional Neural Networks (CNNs) extract spatial features from individual frames, while Bidirectional Long Short-Term Memory (BiLSTM) networks learn the temporal dependencies over time. This combination is particularly effective for dynamic gestures, where the order of movements is crucial to the meaning~\cite{lu2023multimodal}.
    
    \item \textbf{Graph Convolutional Networks (GCNs):} These consider body landmarks as nodes in a graph, allowing the model to learn the structural interconnections between different joints and body parts~\cite{de_Amorim_2019}.
    
    \item \textbf{Transformers:} Utilizing spatial and temporal attention mechanisms, Transformers can capture long-range dependencies across frames and correlations between different visual channels~\cite{ma-etal-2024-multi}.
\end{itemize}

\vspace{-0.5cm}
\subsubsection{Landmark Detection and Efficiency}
In real-time automotive applications, computational efficiency is as vital as accuracy. MediaPipe and OpenPose are the two most frequently cited tools for skeletal feature extraction due to their extensive joint coverage. OpenPose is often prone to temporal jitter and high latency, particularly in occluded environments~\cite{alves2024enhancingbraziliansignlanguage}. In contrast, MediaPipe offers a lightweight alternative that can achieve 5x faster recognition by utilizing an optimized subset of body landmarks~\cite{santos2026properbodylandmarksubset}. Research has shown that a well-chosen landmark subset (e.g., focusing on the upper body and hands) can maintain state-of-the-art accuracy while significantly reducing inference time. While efficient, MediaPipe's performance can degrade significantly when parts of the body are blocked or in poorly lit environments, which often happens inside a car~\cite{Amalfitano2023}. Consequently, there is a need for models that can handle missing hand or body points or combine information from different sensors to deal with visual noise.

\vspace{-0.2cm}
\subsection{The Automotive Domain: Physical and Social Constraints}
The interior of a vehicle presents a highly restricted physical and visual environment for communication. Unlike laboratory settings, where signers have full, unobstructed signing space, the car cabin introduces several ``in-the-wild'' constraints that serve as a starting point for specialized SLR research.

\vspace{-0.2cm}
\subsubsection{Constrained Signing Space and Occlusions}

The physical setup inside a car changes how people sign. Seatbelts cross the body and can obscure important parts of a sign and hinder torso movements.
% Sealbelts across the body limits torso mobility, dashboard, headrests, and tight space all limit arms and elbows movement, affecting sign execution.
The dashboard, headrests, and tight space also limit the movement of the arms and elbows, reducing overall signing space.
\vspace{-0.2cm}
\subsubsection{Environmental Noise and Dynamic Lighting}
Inside a vehicle, visual conditions can change significantly. Lighting can go from bright sunlight to near darkness very quickly. Shadows, reflections from windows, and uneven light can add visual noise, making it harder to clearly see hands and movements. Because of this, sign language recognition systems that work well in lab lighting often struggle to handle these changing real-world conditions~\cite{Amalfitano2023}.
\vspace{-0.2cm}
\subsection{Hardware and Multimodal Sensing}
% In order to make sign language recognition systems more stable, reliable and accurate, researchers employed a multitude of sensors, instead of RGB cameras, to work around the limitations imposed by the vehicular environment.
Because single RGB cameras do not work reliably inside a car, researchers are using multiple sensor types to make sign language recognition systems more stable and accurate.
\vspace{-0.2cm}
\subsubsection{2D vs. 3D Sensing}
While 2D cameras are affordable and widely available in modern vehicles, they lack depth information. Time-of-Flight (TOF)/3D cameras provide 3D skeletal data that can be used to track hand motion trajectories in three dimensions. Depth maps enable the system to segment hands from the background more effectively, especially when the hands are similar in color to the vehicle interior. Depth sensing is critical for sign language because the movement of the hand toward or away from the camera often distinguishes between different signs~\cite{Zhang2022SLR}~\citelanguageresource{3DLEXv1}.
\vspace{-0.2cm}
\subsubsection{Synthetic Data and Motion Capture (MoCap)}
Synthetic data generation has emerged as a key strategy for overcoming the difficulty and high cost of collecting large-scale natural datasets.

Previous studies have successfully utilized motion capture systems to acquire precise three-dimensional point cloud and skeletal data from deaf signers~\cite{kipp2011sign,de2016building,will2018optimized,jedlivcka2020sign,andersen2025designing,de2025neural}. This high-fidelity spatiotemporal data, capturing both manual and non-manual signals, serves as a critical training corpus for neural networks, such as CNNs and RNNs, enabling them to learn the mapping from continuous sign language kinematics to written language translation.

\citet{shterionov2024pipeline} discuss about the types of synthetic work, two of the main trends in synthesis are: (i) the more traditional 3D animation-based which resolves in generating a 3D animated character, commonly referred to as an avatar; and (ii) a video of a virtual human that can be synthesized with generative AI methods based on real human video/image data 

\vspace{-0.3cm}
\section{Motivation for the Set-up}
\label{sec:questions}
\vspace{-0.2cm}
The synthesis of the existing literature reveals the primary motivations for developing our specialized Libras resource. This work distinguishes itself by starting from a stable, fundamental baseline, which is currently missing in the field of automotive sign language recognition.

\begin{itemize}
    \item \textbf{Establishing a Multimodal In-Car Baseline:} There is currently no resource documenting Libras production within the physical and visual constraints of a car cabin. Our setup provides the very first multimodal baseline (RGB + 3D TOF) recorded inside a vehicle. By starting with a simplified, stable environment, we establish the necessary scientific foundation before introducing more complex parameters in future iterations, and it provides the first comparative benchmark between idealized laboratory MoCap and these constrained ``in-the-wild'' conditions.   
    \item \textbf{Comparative Analysis of Constrained Space:} This resource is motivated by the need to quantify the linguistic and physical differences between idealized laboratory signing and signing within the confined space of a car cabin. This dataset allows for a direct comparison between high-precision MoCap and real-world cabin constraints.
    
    \item \textbf{Facilitating Initial Domain Adaptation:} By providing a synchronized real-world baseline and a corresponding synthetic dataset, we enable researchers to begin exploring domain adaptation for cars using expressive 3D avatars.
\end{itemize}

\vspace{-0.5cm}
\subsection{Potential Research Directions}
To address these motivations, the ICSL dataset provides the necessary multimodal data to investigate several key research questions in future studies, such as the following:
%To address these motivations, this dataset aims to answer the following research questions:

\begin{itemize}
    \item \textbf{Camera Placement Optimization:} Determining the optimal camera positions within a vehicle cabin to maximize the visibility of facial expressions and head movements in Libras, despite the spatial constraints.
    \item \textbf{Physical Baseline Comparison:} Analyzing how sign production (hand configuration and path movement) differs when comparing a high-precision laboratory MoCap baseline to the constrained physical space of a vehicle.
    \item \textbf{Multimodal Stability Evaluation:} Assess to what extent multimodal sensor fusion improves the stability of landmark detection in a confined interior compared to traditional monocular systems.
    \item \textbf{Perspective Gap Assessment:} Evaluating whether signing avatars, generated from idealized MoCap, can effectively simulate the non-frontal viewing angles required for in-car Human-Machine Interfaces (HMI).
    \item \textbf{Linguistic Adaptation Analysis:} Investigating how physical restrictions, such as the car seat and seatbelt, impact the duration and spatial volume of signs compared to laboratory-recorded data.
    \item \textbf{Model Robustness Testing:} Quantifying the performance drop when models trained on stable in-car data are tested against other driving-related micro-gestures or environmental noise.
\end{itemize}

By establishing this baseline, we provide the scientific community with a foundation for building inclusive, real-time assistive technologies that bridge the communication gap in shared mobility services.

\vspace{-0.2cm}
\section{Corpus Design}
The design of the In-Car Sign Language (ICSL) corpus is driven by the need for accessible passenger-driver communication within shared mobility services. Unlike general-purpose sign language datasets, this resource focuses exclusively on the constrained signing space of vehicle interiors to analyze how physical constraints impact LIBRAS language. 

\vspace{-0.2cm}
\subsection{Scenario Selection and Use Cases}
%We have defined 127 essential communication scenarios tailored for the taxi ride-hailing domain. These use cases are categorized into five functional groups:
In collaboration with our Deaf researchers and professional Libras interpreters, we defined 1,344 essential communication scenarios for shared mobility services.
%These use cases, 
To date, we have recorded 127 essential cases that reflect the immediate needs of DHH passengers. These are categorized into five functional groups:

\begin{itemize}
    \vspace{-0.1cm}
    \item \textbf{Navigation and Routing}
    \vspace{-0.2cm}
    \begin{itemize}
        \item ``Qual o destino?'' (Where to?)
        \item ``Melhor evitar o centro.'' (It is best to avoid downtown.)
        \item ``Já passamos o ponto.'' (We are past my drop-off.)
        \item etc.
    \end{itemize}
    \item \textbf{Safety and Regulations}
    \vspace{-0.2cm}
    \begin{itemize}
        \item ``Reduzindo por segurança.'' (Reducing for safety reasons.)
        \item ``Pode ir devagar?'' (Can you slow down?)
        \item etc.
    \end{itemize}
    \item \textbf{Passenger Comfort and Interaction}
    \vspace{-0.2cm}
    \begin{itemize}
        \item ``Preciso descer urgente.'' (I need to stop urgently.)
        \item ``Posso ligar o ar-condicionado?'' (May I turn on the air conditioning?)
        \item etc.
    \end{itemize}
    \item \textbf{Assistance}
    \vspace{-0.2cm}
    \begin{itemize}
        \item ``Pode ajudar com a cadeira de rodas?'' (Can you help with the wheelchair?)
        \item ``Tem troco para R\$ 25,00?'' (Do you have change for R\$ 25,00?)
        \item ``Tenho uma mala pesada.'' (My luggage is heavy.)
        \item etc.
    \end{itemize}
    \item \textbf{Alphabet Spelling}
    \vspace{-0.2cm}
    \begin{itemize}
        \item A, B, C, etc.
    \end{itemize}
\end{itemize}

\vspace{-0.3cm}
\subsubsection{Subject Recruitment and Diversity}

\begin{figure}[!htb]
    \centering
    \subfloat[]{\includegraphics[width=0.155\textwidth]{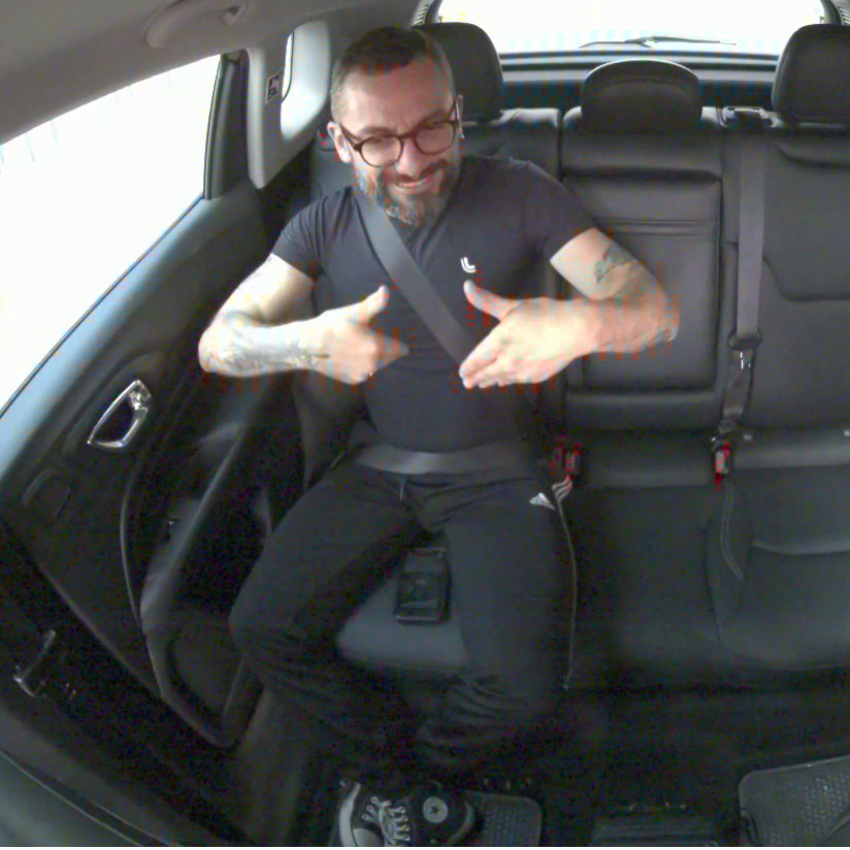}} \hspace*{0.01cm}
    \subfloat[]{\includegraphics[width=0.155\textwidth]{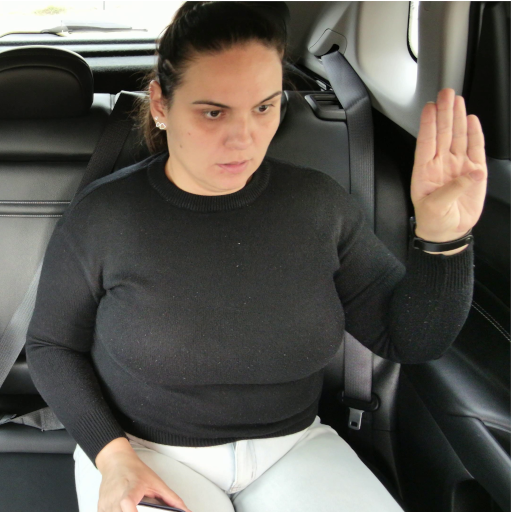}} \hspace*{0.01cm}
    \subfloat[]{\includegraphics[width=0.155\textwidth]{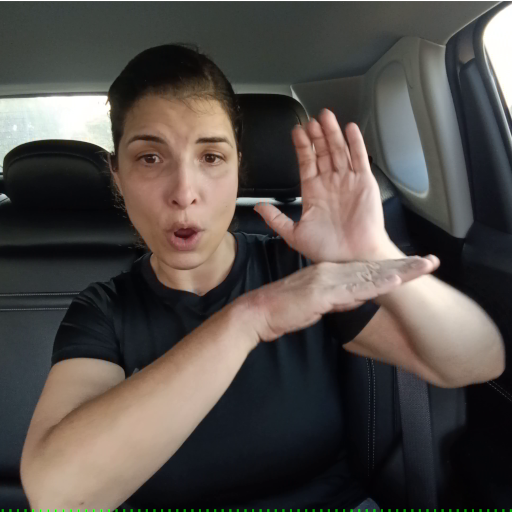}}
    \caption{Illustration of the three signers and challenges within vehicle cabin: (a) FLIR camera, (b) ToF camera, and (c) recording tablet point of views.}
    \label{fig:multimodal_views}
\end{figure}

The corpus was recorded by one hearing and two deaf signers, all proficient in Libras, which were instructed to dress as usual for the capture session. They also served as annotators to our data. To ensure that our data captures a wide range of physical constraints, recordings were conducted across three different vehicle models with varying cabin geometries: a Jeep Compass, a Nissan March, and a Renault Megane. All sessions were performed under clear daylight conditions to minimize extreme external lighting noise while preserving the natural dynamic shadows of the vehicle interior as seen in Figure~\ref{fig:multimodal_views}.

\vspace{-0.2cm}

\vspace{-0.5cm}
\section{Data Acquisition and Technical Setup}
Our data collection followed a two-stage approach. We first established a laboratory baseline using high-precision motion capture to define the ``gold standard'' for each sign. This was followed by an acquisition phase inside a vehicle cabin, using a multimodal sensor setup to capture the natural constraints for in-car communication.
\vspace{-0.2cm}
\subsection{Laboratory MoCap Baseline}
Our ICSL MoCap corpus was built to simulate in-vehicle body positioning and its effects on Libras sign production. To do so, we employed several sensors, including RGB cameras and a motion capture suit (MoCap), to record joint movements with detail and precision during sign language communication.
The setup made use of several professionals, including: 
\begin{itemize}
    \item Two Libras signers who were fluent in Libras and performed the signs.
    \item One system operator to up-keep the quality of the MoCap data being recorded.
    \item Two hardware operators to ensure frame synchronization and camera operation, and lastly.
    \item One content facilitator to assist with translation and language production evaluation.
\end{itemize}
%(i) two Libras signers who were fluent in Libras and performed the signs, (ii) one system operator to up-keep the quality of the MoCap data being recorded, (iii) two hardware operators to ensure frame synchronization and camera operation, and lastly (iv) one content facilitator to assist with translation and language production evaluation.

\begin{figure}[!htb]
    \centering
    \subfloat[]{\includegraphics[width=0.155\textwidth]{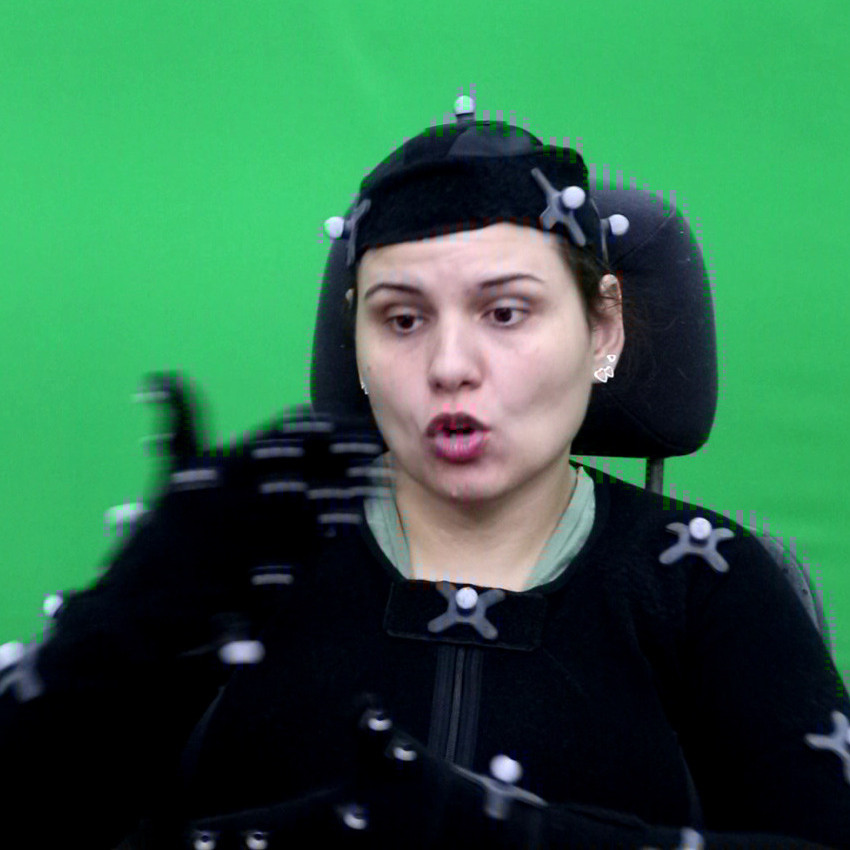}\label{sfig:mocap_suit}} \hspace*{0.01cm}
    \subfloat[]{\includegraphics[width=0.155\textwidth]{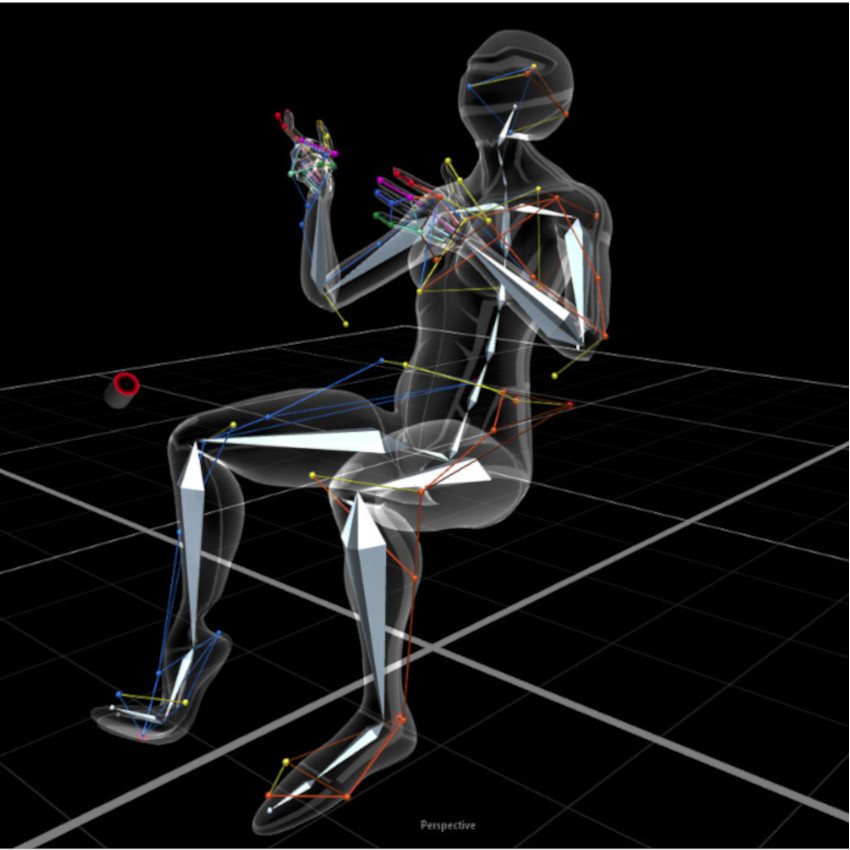}\label{sfig:mocap_skeleton}} \hspace*{0.01cm}
    \subfloat[]{\includegraphics[width=0.155\textwidth]{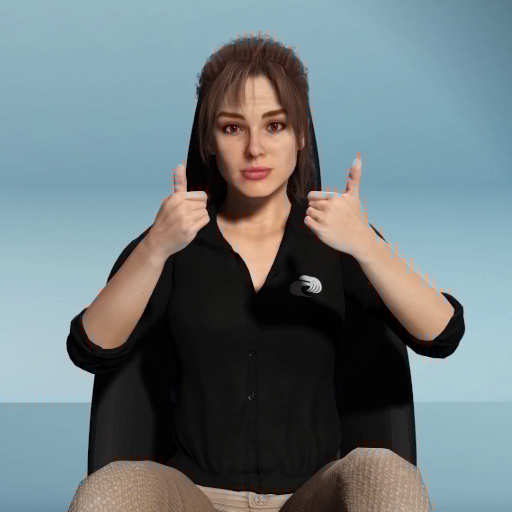}\label{sfig:clara}}
    \caption{MoCap setup illustrating the (a) capturing suit, (b) the markerset and skeleton from the Vicon tracking software, and (c) the rendered avatar (Clara).}
    \label{fig:mocap_setup}
\end{figure}

Regarding our frame composition for this task, the laboratory has a green-screen background and a mock-up car seat. The Libras' signers are recorded while seated, wearing a special MoCap suit that covers most of their bodies. The suit is made from black cloth and allows the attachment of reflective beads (or spherical markers) with velcro, which are detectable by the infrared cameras as seen in Figure~\ref{sfig:mocap_suit}. In total, there are 73 beads distributed across the main signer's joints.

We also distributed fourteen Vicon~\cite{ViconValkyrie} cameras radially around the signer. These cameras are equipped with near-IR sensor that detect the spherical markers attached to the MoCap suit. Later on the markers will be digitalized to move the character skeleton in the capture software as illustrated in Figure~\ref{sfig:mocap_skeleton}, which in turn will animate the avatar as shown in Figure~\ref{sfig:clara}. Moreover, two Canon EOS T6i Rebel~\cite{CanonT6i} are also employed for RGB capture, alongside the FLIR 2D~\cite{TeledyneFireflyS}, and the Vicon Vue cameras.

\begin{comment}
\vspace{-0.5cm}
\begin{figure}[ht]
    \centering
    \subfloat[]{\includegraphics[width=0.24\textwidth]{Images/mocap_suit}\label{sfig:mocap_suit}} \hspace*{0.01cm}
    \subfloat[]{\includegraphics[width=0.24\textwidth]{Images/mocap_skeleton}\label{sfig:mocap_skeleton}}\\
    \subfloat[]{\includegraphics[width=0.24\textwidth]{Images/clara}}
    \caption{MoCap setup illustrating the (a) capturing suit, (b) the character mesh and skeleton from the Vicon tracking software, and (c) the rendered avatar (Clara).}
    \label{fig:mocap_setup}
\end{figure}
\end{comment}

\vspace{-0.2cm}
\subsection{In-Car Data Acquisition}
The in-car component of the ICSL corpus was designed to capture the linguistic and environmental challenges of a transport system setting. We employed a multi-sensor approach to provide both standard video and high-fidelity 3D spatial data.

\vspace{-0.2cm}
\subsubsection{Hardware and Sensor Setup}
\begin{figure}[!htb]
    \centering
    \subfloat[]{\includegraphics[width=0.155\textwidth]{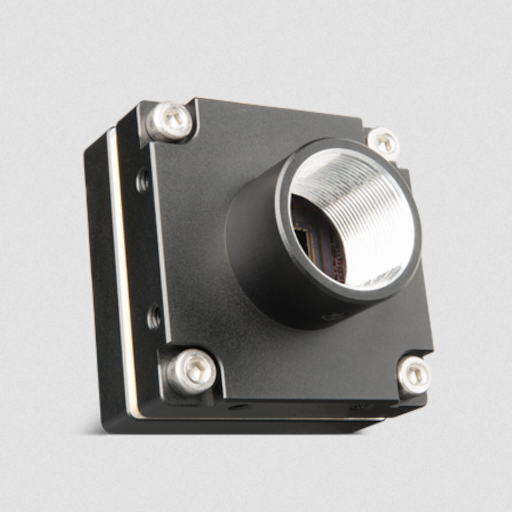}} \hspace*{0.01cm}
    \subfloat[]{\includegraphics[width=0.155\textwidth]{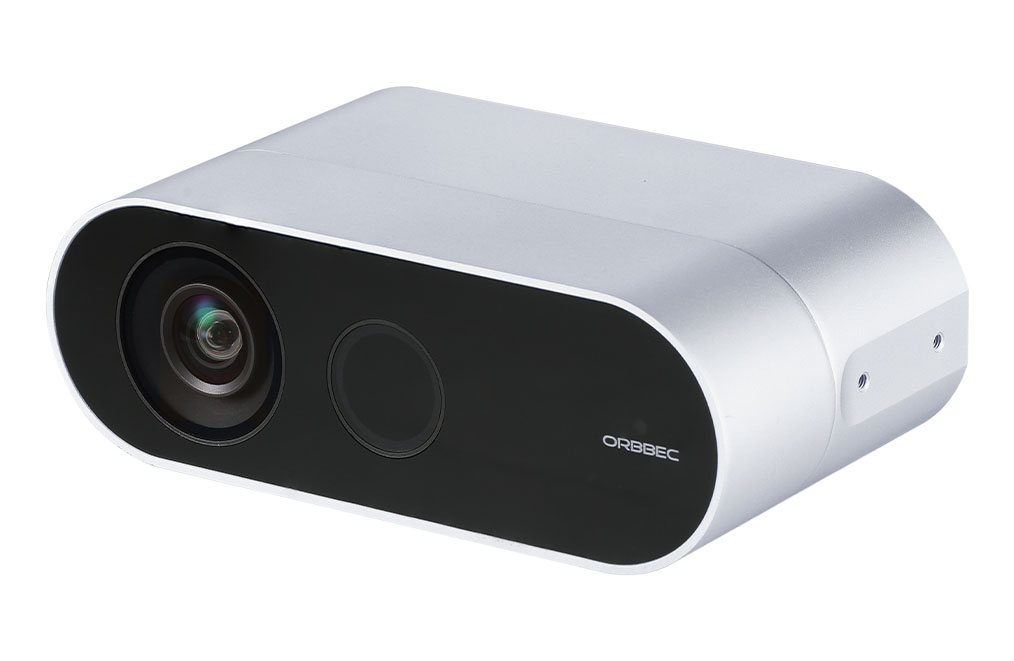}} \hspace*{0.01cm}
    \subfloat[]{\includegraphics[width=0.155\textwidth]{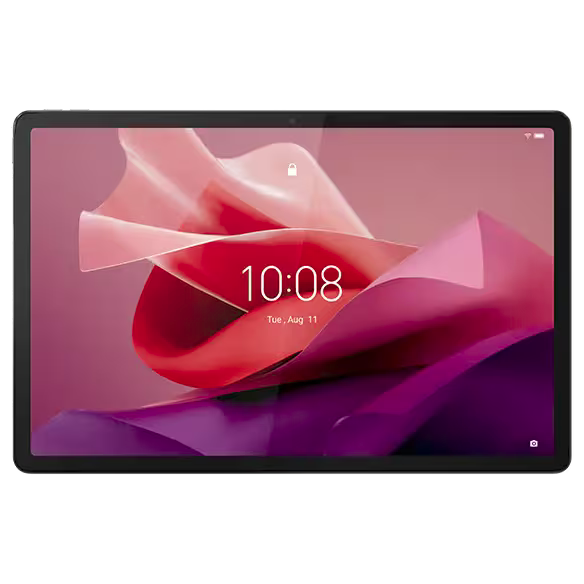}}
    \caption{The multi-modal sensor suite including (a) the FLIR 2D camera, (b) Orbbec ToF sensor, and (c) the Lenovo Tablet.}
    \label{fig:sensors}
\end{figure}

\begin{figure}[!htb]
    \centering
    \subfloat[]{\includegraphics[width=0.40\textwidth]{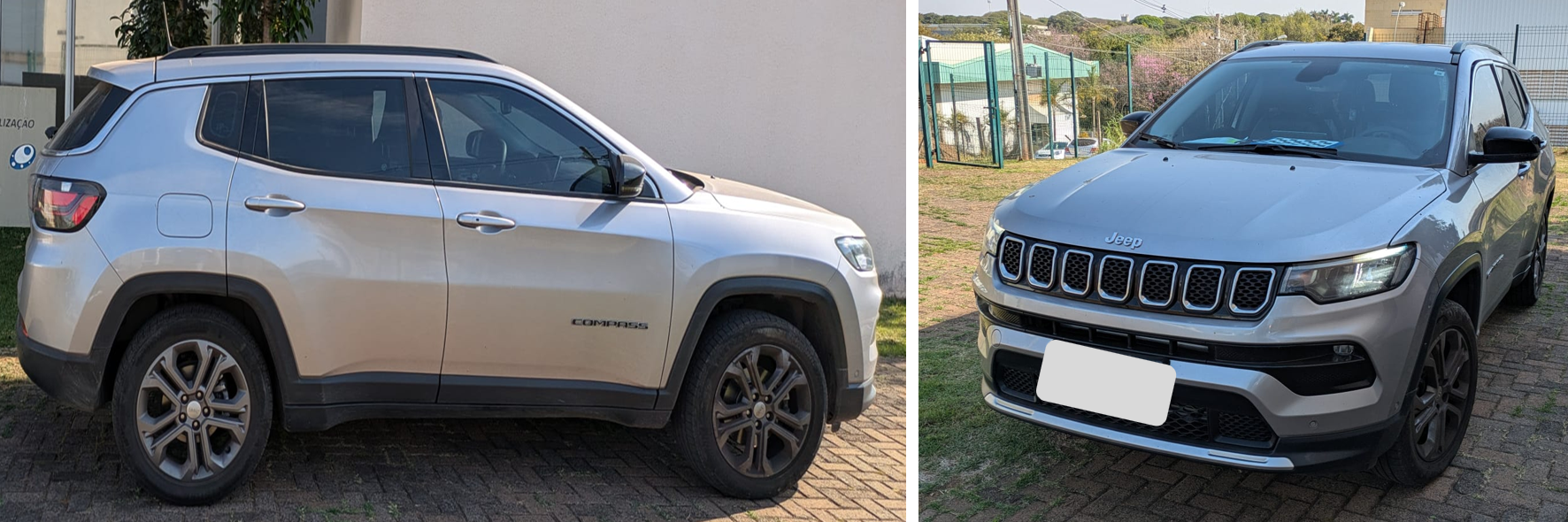}\label{sfig:jeep_setup_1}} \\
    \subfloat[]{\includegraphics[width=0.40\textwidth]{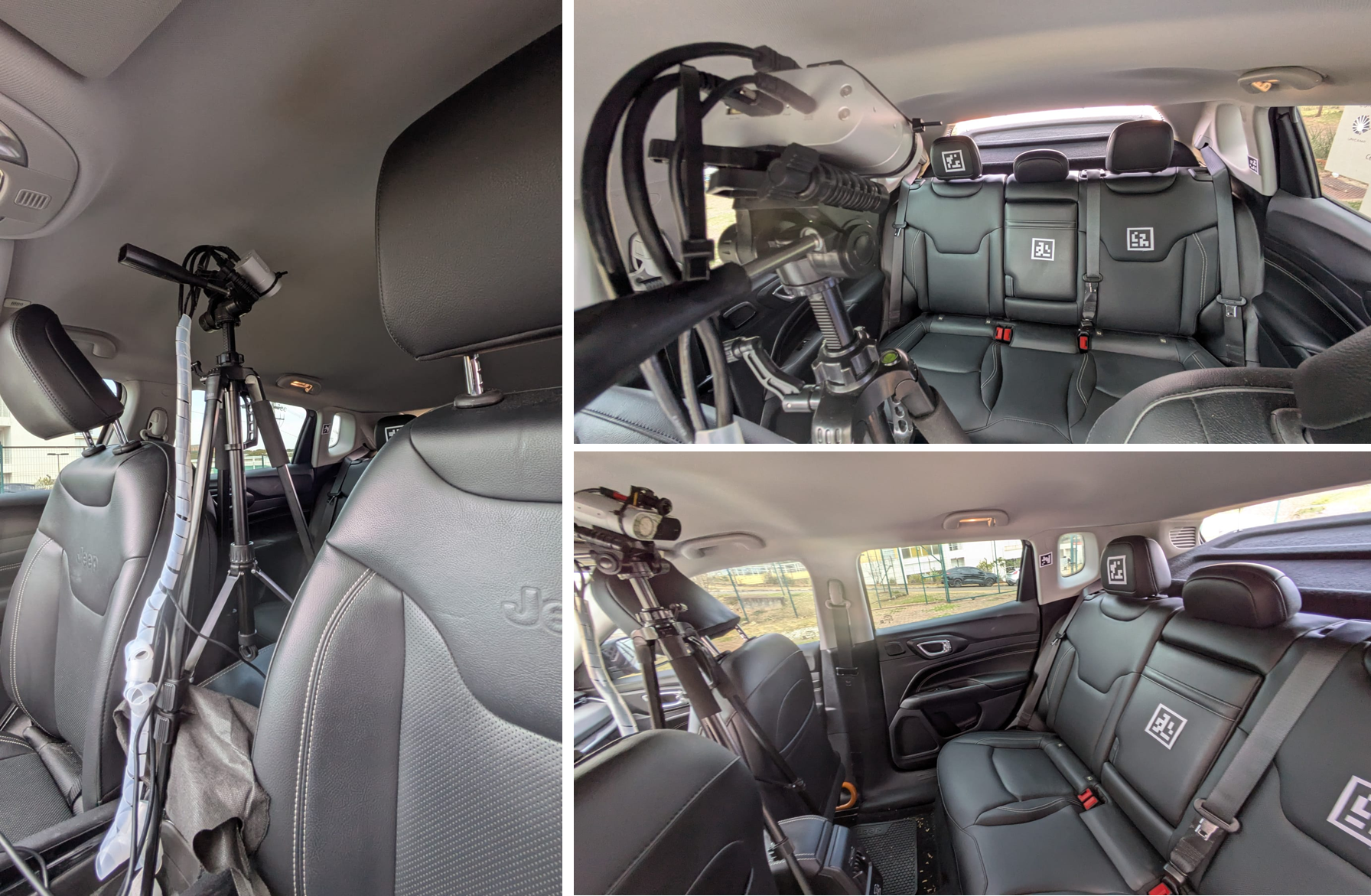}\label{sfig:jeep_setup_2}}\\
    \caption{(a) test vehicle, and (b) full technical installation inside the Jeep Compass cabin, illustrating sensor placement and the passenger signing space.}
    \label{fig:setup}
\end{figure}

We utilized multi-modal sensors as seen in Figure~\ref{fig:sensors} and the setup as seen in Figure~\ref{fig:setup}b was deployed across three distinct test vehicles: a Jeep Compass(illustrated in Figure~\ref{fig:setup}a), a Nissan March, and a Renault Megane, providing variability in cabin dimensions, window size, lightning, color, texture, seat shape, and interior geometry.

\begin{comment}
\begin{figure}[ht]
    \centering
    \subfloat[]{\includegraphics[width=0.2\textwidth]{Images/depth}} \hspace*{0.01cm}
    \subfloat[]{\includegraphics[width=0.2\textwidth]{Images/ir}}
    \caption{Multimodal perspectives of Libras communication within a vehicle cabin, illustrating the two sensors: (a) a ToF depth sensor using a color scale, ranging from blue for nearby objects to green for distant ones, with dark blue indicating areas where no reading was captured, and (b) a ToF IR sensor displayed in grayscale, with white representing surfaces that reflect infrared light and black representing those that do not.}
    \label{fig:multimodal_sensors}
\end{figure}
\end{comment}
\vspace{-0.5cm}

\begin{figure}[!htb]
    \centering
    \subfloat[]{\includegraphics[width=0.18\textwidth]{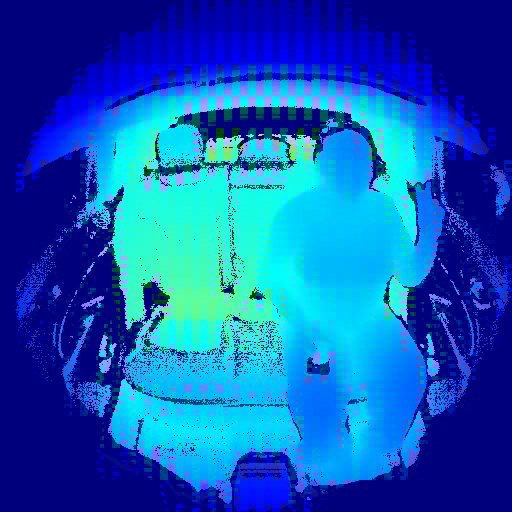}} \hspace*{0.2cm}
    \subfloat[]{\includegraphics[width=0.18\textwidth]{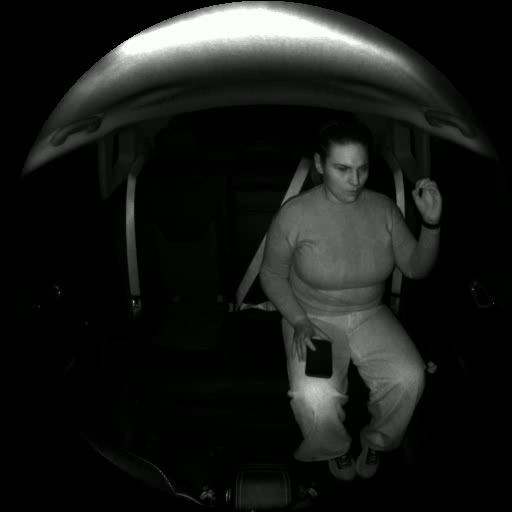}} \\
    \vspace{-0.3cm}
    \subfloat[]{\includegraphics[width=0.18\textwidth]{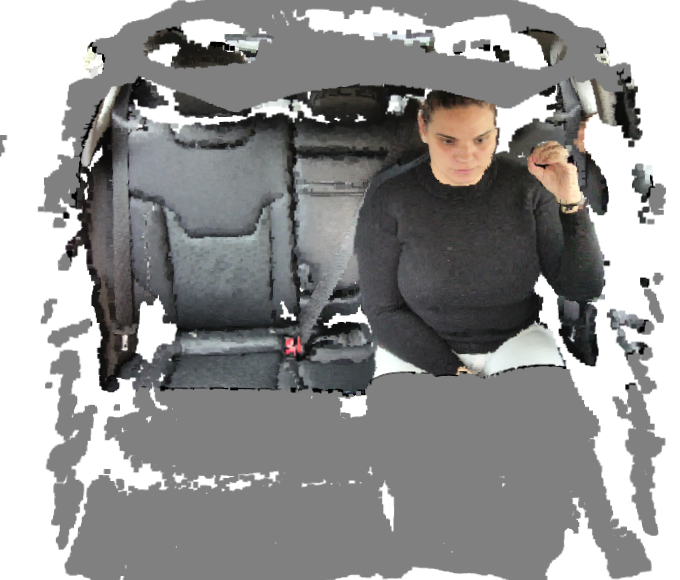}} \hspace*{0.2cm}
    \subfloat[]{\includegraphics[width=0.18\textwidth]{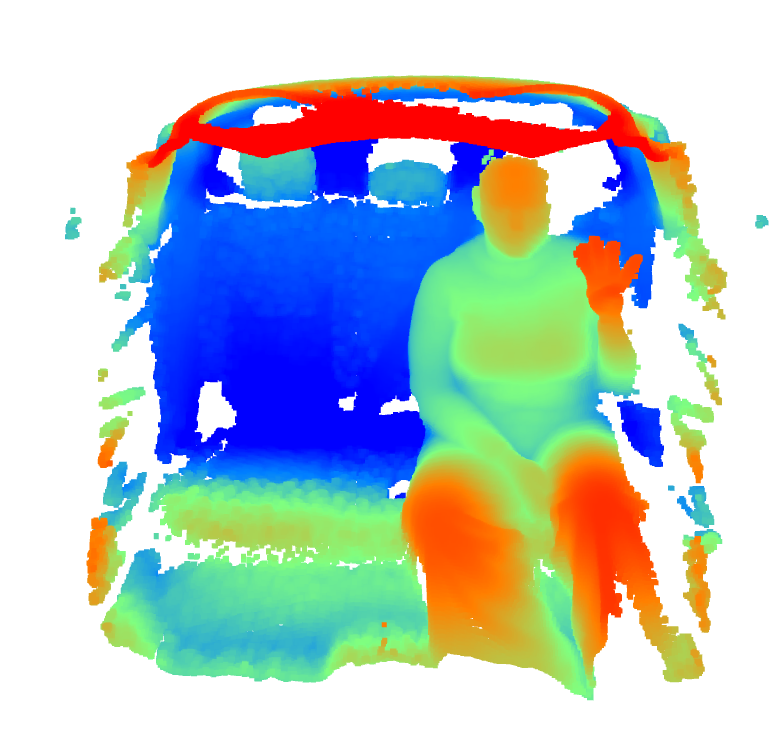}}\\
    \caption{Multimodal perspectives of Libras communication within a vehicle cabin, illustrating the two sensors: (a) ToF depth image shown in color, where blue indicates closer objects and green indicates farther objects. Dark blue areas mean no depth was measured, (b) ToF infrared (IR) image shown in grayscale, where white represents strong IR reflection and black represents weak or no reflection, (c) ToF RGB point cloud, where each 3D point contains spatial coordinates $(X, Y, Z)$ and its corresponding color (R, G, B), and (d) ToF Depth point cloud, where each 3D point contains spatial coordinates $(X, Y, Z)$ derived only from depth measurements, without color information.}
    \label{fig:multimodal_sensors}
\end{figure}

\vspace{0.35cm}
\begin{itemize}
    \item \textbf{2D Imaging:} A FLIR Blackfly S~\cite{TeledyneFireflyS} was utilized to capture the primary 2D visual stream at a resolution of $1440 \times 1080$, operating at 60 fps as seen in Figure~\ref{fig:sensors}a. To accommodate the confined vehicle interior, we utilized fisheye lenses to maximize the field of view as seen in Figure~\ref{fig:multimodal_views}a.
    \item \textbf{3D Time-of-Flight (ToF):} An Orbbec Femto Bolt~\cite{OrbbecFemtoBolt2026} was utilized to capture spatial information as seen in Figure~\ref{fig:sensors}b. The sensor was configured to capture synchronized RGB, Depth, Infrared (IR), and Point Cloud streams at a resolution of $1024 \times 1024$ pixels, operating at 30 fps, allowing for precise tracking of hand movements in 3D space as seen in Figures~\ref{fig:multimodal_views}b and~\ref{fig:multimodal_sensors}a-d. 
    \item \textbf{Recording Tablet:} A Lenovo Tab P12~\cite{LenovoTabP12} was utilized as an additional perspective, as seen in Figure~\ref{fig:sensors}c. Positioned behind the front-seat headrest at a resolution of up to $3840 \times 2160$ pixels, it recorded the signer from an elevated, centered viewpoint, providing a third data stream to supplement the mounted sensors as seen in Figure~\ref{fig:multimodal_views}c.
    \item \textbf{Prompting Interface (Mobile Phones):} In addition to the main sensors, mobile phones were positioned within the signer's line of sight specifically to display the selected use cases. These served as a remote prompting system to guide the signer through the recording sequence.
\end{itemize}

\vspace{-0.5cm}
\subsection{Acquisition Protocol and Workflow}
\begin{figure}[ht]
    \centering
    \includegraphics[width=8cm, height=5cm]{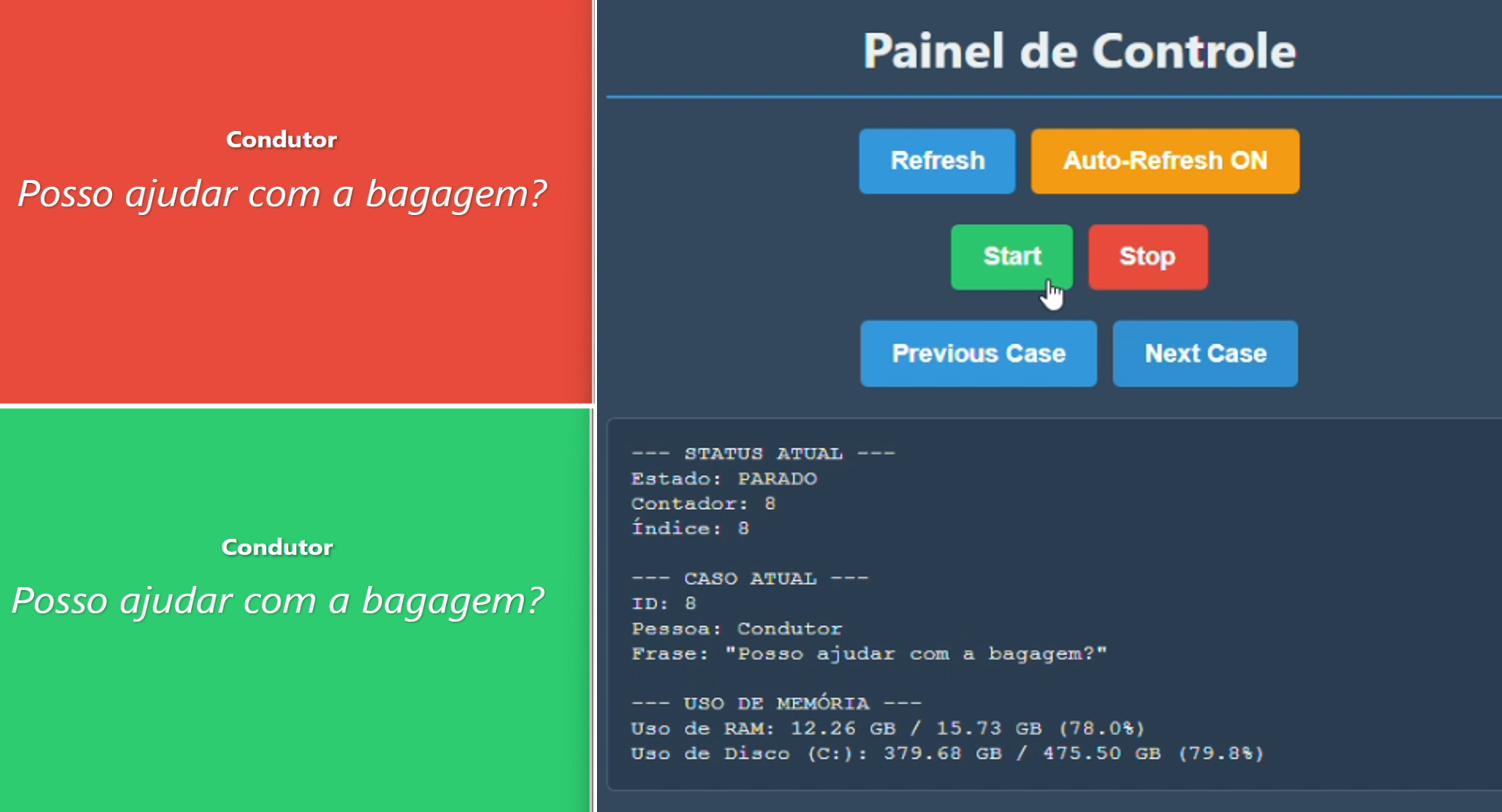}
    \caption{Web-based remote control interface for Start and Stop signaling.}
    \label{fig:webView}
\end{figure}

The recording session followed a strictly controlled protocol managed via a custom web-based control panel, as seen in Figure~\ref{fig:webView}. To ensure the signer remained focused and in the correct posture, we utilized a remote signaling system:

\begin{itemize}
    \item \textbf{Instruction Delivery:} Use cases were published on the mobile phone screens from the central Control Panel.
    \item \textbf{Start/Stop Signals:} The operator sent a ``Start'' signal (green) to the phone to indicate which sign to perform. Once the gesture was completed, a ``Stop'' signal (red) was sent, and the system automatically switched to the ``Next Case'' for the signer.
\end{itemize}

\vspace{-0.2cm}

\begin{comment}
\subsubsection{Corpus Statistics}
The current collection phase has resulted in a substantial high-resolutioin source:
\begin{itemize}
    \item \textbf{FLIR 2D Frames:} 737948
    \item \textbf{ToF 3D/RGB/IR/Depth Frames:} 368974
    \item \textbf{Tablet Recording Frames:} 428162
\end{itemize}
Totaling over 1.5 million frames, this resource provides the necessary volume and diversity for evaluating ``in-the-wild'' sign language recognition models within the automotive domain. \textcolor{red}{need to add the sync. frames count}
\end{comment}

% \subsection{MoCap Recording for In-Car usecases}

% {\color{blue}pictures from the avatar was rendered for these usecases 
% how many frames in the lab setup}

% i think we can eliminate this section, since the phrases are the same as inside the car

\vspace{-0.2cm}
\section{Corpus Analysis and Limitations}
\label{sec:limitations}
\vspace{-0.2cm}
Our dataset represents a 
%deliberate
first step toward a comprehensive in-vehicle Libras resource. To ensure the scientific validity of our comparative analysis, we chose to establish a solid 
% multimodal 
baseline under stable, controlled conditions before introducing more dynamic environmental variables. This approach allows us to demonstrate the causal relationship between the physical layout of the car and signing production without the interference of visual noise.
%The current corpus provides synchronized RGB and 3D TOF data (Point Cloud, Depth and IR) for specific passenger-driver communication scenarios. 
By focusing on these idealized conditions, we provide a ``ground truth'' reference that enables the evaluation of technical benchmarks in a confined space. This is critical because sign language representations must reflect 3D awareness to handle the orientations and spatial trajectories inherent in real-world cabins. 

The current iteration of the dataset is built upon a theoretical vocabulary of 1,344 Portuguese context-specific phrases for ride-sharing applications. To date, a subset of 127 phrases has been successfully recorded in video, with plans to complete the remaining recordings in the coming months. The data collection and annotation methodology relies on a cohort of three participants (two deaf, one hearing) who perform both the signing and the ongoing data annotation.

A significant current limitation is the lack of dynamic lighting conditions, such as sun glare, tunnel transitions, or nighttime driving noise. While these factors are vital for ``in-the-wild'' robustness, in this initial phase, we theorize that models pre-trained on normal illumination data serve as a superior foundation for subsequent fine-tuning in low-light scenarios. As an early-stage resource within Project UNITY, the dataset currently features a limited number of signers, two deaf signer and one hearing. We acknowledge this as a potential source of signer bias that could limit the generalization of initial AI models. While we capture stationary occlusions (seatbelts, headrests), the dataset does not yet account for dynamic occlusions caused by vehicle movement or secondary passenger interactions.

By acknowledging these constraints, we frame this work not as a final solution but as the essential foundation required to guide our future, more complex data-collection efforts in the automotive sign language domain.

\begin{comment}

\vspace{-0.3cm}
\section{The UNITY Bidirectional Communication Framework}
\vspace{-0.2cm}
The goal of Project UNITY is to bridge the communication gap in automotive environments through a real-time, bidirectional translation system as seen in Figure~\ref{fig:architecture_top}. Unlike previous one-way approaches that focus solely on translating driver instructions to the passenger, our proposed architecture facilitates a complete conversational loop between spoken language used by the driver and sign language (Libras/LSF) used by the Deaf or Hard-of-Hearing (DHH) passenger.

\end{comment}

\vspace{-0.2cm}
\section{Conclusion}
\vspace{-0.2cm}
This paper introduces the In-Car Sign Language (ICSL) dataset, a novel multimodal resource designed to 
%address the critical gap in 
investigate the technical challenges of sign language recognition across shared mobility services. By focusing specifically on the constrained signing environment of vehicle interiors, we have established the first specialized dataset that systematically documents how Brazilian Sign Language (Libras) is produced within the physical and visual limitations of car cabins. The corpus comprises over 1.5 million synchronized frames across multiple modalities,
%including high-precision laboratory MoCap, real-world in-car RGB, and 3D Time-of-Flight recordings
providing researchers with a comprehensive foundation for investigating sign language recognition in real mobile settings.

%Our contribution is split into two parts. First, we have created a comparative benchmark that enables direct analysis between idealized laboratory signing and the constrained conditions of actual vehicle interiors, addressing the research questions posed in Section~\ref{sec:questions} regarding physical adaptations, multimodal stability, and linguistic modifications. Second, by documenting 127 essential passenger-driver communication scenarios across three distinct vehicle models, we have ensured that this resource reflects real-world accessibility needs while capturing the variability introduced by different cabin geometries.

Our contribution is split into two parts. First, a comparative benchmark between laboratory and in-car signing that addresses potential research directions in Section~\ref{sec:questions} focused on physical, multimodal, and linguistic adaptations. Second, we provide 127 communication scenarios recorded across three vehicle models, reflecting real-world accessibility needs within diverse cabin geometries.

The limitations acknowledged in Section~\ref{sec:limitations} including controlled lighting conditions, and signer diversity, this work establishes a necessary foundation for robust ``in-the-wild'' SLR systems. As a component of Project UNITY, the ICSL corpus provides an empirical baseline for assistive technologies aiming to reduce communication barriers between hearing drivers and DHH passengers. Available upon request, this resource supports the development of inclusive transportation systems where language barriers no longer hinder DHH access and dignity.

\vspace{-0.25cm}
\section{Future Work}
\vspace{-0.2cm}
Moving forward, our next step is to introduce dynamic environmental parameters. We plan to record subsequent datasets under varied lighting conditions. Additionally, future iterations will move from a stationary vehicle to active driving environments to capture dynamic occlusions and noise caused by vehicle vibration, which are known bottlenecks for real-time landmark tracking. Although our dataset focuses on sentence-based communication scenarios, we aim to expand the ICSL corpus to include continuous Libras conversations between passengers, more complex phrases, and a broader range of subjects.

% {\color{blue}other transportation modals (inter-ciity bus, in-city bus, train, plane, ship, vans, metro, bondinho), more people, more complex phrases, moving vehicle, more cars, can oyu write abt the future work wrt Mocap, how can we blend it here?}

\vspace{-0.2cm}
\section{Acknowledgement}
\vspace{-0.2cm}
% TODO Missing grants, double check if there is a FAPESP grant specific to UNITY and a grant from germany.
%{\color{red}Probably there is a second FAPESP grant for the project UNITY, and a grant from Germany. }
This project is partially financed by the São Paulo Research Foundation (FAPESP), grants \#2024/23068-4 and \#2024/00914-7, the Brazilian Federal Agency for Support and Evaluation of Graduate Education (CAPES), grant \#88887.091672/2014-01, National Council for Scientific and Technological Development (CNPq), grant \#458691/2013-5, and FINEP, grant \#2778/20. The authors would also like to thank the Verkehrsverbund Großraum Ingolstadt (VGI) for their support through the VGI newMIND project.
%, which is funded by the German Federal Ministry for Digital and Transport (BMDV).

\vspace{-0.2cm}
\section{Ethical Considerations}
\vspace{-0.2cm}
The Libras signers in this project are not just participants, they are paid team members and co-authors of this study. Deaf and hard-of-hearing researchers were involved in every stage of the project, from designing the data collection process to writing the paper. This helps us make sure that the linguistic details of Libras are treated with care and that the technology we develop truly reflects the needs of the community. All team members gave their informed consent for the use of their images and biometric data in the corpus. Since the data was created by the research team specifically for this resource, the project follows the institutional ethical guidelines for collaborative research. To request access to the corpus or for further technical inquiries regarding the acquisition software, annotation tools, or the Project UNITY framework, please contact José Mario De Martino at \url{martino@unicamp.br}.

%{\color{blue}All Libras signers involved in the project, Deaf or otherwise, are paid project members and coauthors of the paper. No external participants took part in the study; therefore, formal ethical approval was not required.}

\section{Bibliographical References}\label{sec:reference}
\vspace{-1cm}
\bibliographystyle{lrec2026-natbib}
\bibliography{references}

\section{Language Resource References}
\label{lr:ref}
\vspace{-1cm}
\bibliographystylelanguageresource{lrec2026-natbib}
\bibliographylanguageresource{languageresource}

\end{document}